\documentclass[10pt,journal,compsoc]{IEEEtran}

\ifCLASSOPTIONcompsoc
  \usepackage[nocompress]{cite}
\else
  \usepackage{cite}
\fi
\ifCLASSINFOpdf
\else
\fi
\usepackage{times}
\usepackage{epsfig}
\usepackage{graphicx}
\usepackage{amsmath}
\usepackage{amssymb}
\usepackage{multirow} 
\usepackage[pagebackref=true,breaklinks=true,letterpaper=true,colorlinks,bookmarks=false]{hyperref}
\usepackage[nolines,nowrite]{minorrevision}

\makeatletter
\def\paragraph{\@startsection{paragraph}{4}{\z@}{1.5ex plus 1.5ex minus 0.5ex}%
{0ex}{\normalfont\normalsize\bfseries}}
\def\@IEEEsectpunct{.\ \,}
\makeatother

\begin{document}
%
\title{Analysing domain shift factors between videos and images for object detection}
%
%
%
%

\author{Vicky~Kalogeiton,
        Vittorio~Ferrari,
        and~Cordelia~Schmid
\IEEEcompsocitemizethanks{
\IEEEcompsocthanksitem V. Kalogeiton is with the CALVIN team at the University of Edinburgh and with the LEAR team at INRIA Grenoble.\protect\\
E-mail: vicky.kalogeiton@ed.ac.uk
\IEEEcompsocthanksitem V. Ferrari is with the CALVIN team at the University of Edinburgh.\protect\\
E-mail: vferrari@staffmail.ed.ac.uk
\IEEEcompsocthanksitem C. Schmid is with the LEAR team at INRIA Grenoble.\protect\\
E-mail: cordelia.schmid@inria.fr
}
}

\IEEEtitleabstractindextext{%
\begin{abstract}
Object detection is one of the most important challenges in computer
vision. Object detectors are usually trained on bounding-boxes from
still images. Recently, video has been used as an alternative source
of data. Yet, for a given test domain (image or video), the performance of the detector depends on the domain it was trained on.
In this paper, we examine the reasons behind this performance gap. 
We define and evaluate different domain shift factors: spatial location accuracy, appearance
diversity, image quality and aspect distribution.
We examine the impact of these factors by comparing performance before and after factoring them out. 
The results show that all four factors affect the performance of the detectors and their combined effect explains nearly the whole performance gap.
\end{abstract}

}

\maketitle

\IEEEdisplaynontitleabstractindextext

%
\IEEEpeerreviewmaketitle

\ifCLASSOPTIONcompsoc
\IEEEraisesectionheading{\section{Introduction}\label{sec:introduction}}
\else
\section{Introduction}
\label{sec:introduction}
\fi

%
%
%
%

\IEEEPARstart{O}{bject} class detection is a central problem in computer vision.
Object detectors are usually trained on
still images. Traditionally, training an object detector requires
gathering a large, diverse set of still images, in which objects are
manually annotated by a bounding-box~
\cite{Dalal05:thomas,felzenszwalb10pami,girshick14cvpr,MalisiewiczICCV11,Uijlings13,Viola01,wang13iccv}
This manual annotation task can be very time consuming and expensive.
This led the computer vision community to trying to reduce the
amount of supervision necessary to train an object detector, typically down to just a binary label indicating whether the object is present~\cite{Chum07a,cinbis2014cvpr,deselaers12ijcv,Fergus03:thomas,pandey11iccv,siva13cvpr,siva11iccv,siva12eccv,song14icml,Wang2014}.
However, learning a detector with weakly supervision is very challenging and current performance is still well below fully supervised methods~\cite{cinbis2014cvpr,song14icml,Wang2014}. 

Video can be used as an alternative rich source of data. As opposed to
still images, video provides several advantages:
(a)~motion enables to automatically segment the object from the background~\cite{Papazoglou_2013_ICCV}, replacing the need for manually drawing bounding-boxes;
(b)~a single video often shows multiple views of an object; and
(c)~multiple deformation and articulation states (e.g. for animal classes).
Recent work~\cite{Kim14cvpr,Leistner11,prest12cvpr,sharma2013cvpr,tang2012nips,Tang2013} 
started to exploit both sources of data for object detection, by transferring
information extracted from the video domain to the still images domain, or vice versa.  
Hence, these works operate in a domain adaptation setting~\cite{Pan10}.

Several approaches for object detection exist, in which the source domain is video and the target domain is still images \cite{Leistner11,prest12cvpr,Tang2013}.
Leistner et al. \cite{Leistner11} use patches extracted from unlabelled videos to regularize the learning of a random forest detector on still images.
%
%
Prest et al.~\cite{prest12cvpr} and Tang et al.~\cite{Tang2013} present weakly supervised techniques for automatically annotating spatio-temporal segments on objects in videos tagged as containing a given class. These are then used to train object detectors. However, the experiments in~\cite{prest12cvpr} show that training object detectors on still images outperforms training on video frames. 

Other works such as~\cite{sharma2013cvpr,tang2012nips} use still images as source domain and
video frames as target domain.
Tang et al. \cite{tang2012nips} introduce a self-paced domain
adaptation algorithm to iteratively adapt an object detector
from labeled images to unlabeled videos. 
Sharma and Nevatia~\cite{sharma2013cvpr} propose an on-line adaptation
method, which adapts a detector trained off-line on images to a test
video. 
They 
show that the performance of the detector on videos can be significantly improved
by this adaptation, as the initial image training samples and the test video samples can be very
different. 

The above works show that when testing on a target domain,
there is a significant performance gap between training on this domain or
on a different one. This is due to the different
nature of the two domains. 
In this paper, we explore the differences between still images and video frames for training and testing an object detector. We consider several domain shift factors that make still images different from video frames. 
To the best of our knowledge, we are the first to analyze with a structured protocol such domain shift factors so as to reveal the source of the performance gap.

We carry out our investigation on two image-video dataset pairs.
The first pair is PASCAL VOC 07~\cite{pascal07:thomas} (images) and YouTube-Objects~\cite{prest12cvpr} (video). Both datasets in the second pair come from ILSVRC 2015 \cite{ILSVRC15,ImgNet2015}, i.e. from the 'object detection in images' and 'object detection in video' tracks of the challenge, respectively.
We identify and analyse five kinds of domain shift factors that make still images different from video frames (sec.~\ref{sec:factors}).
%
The first is the {\em spatial location accuracy} of the training samples (sec.~\ref{sub:SLA}). As most previous experiments on training detectors from video were done in a weakly supervised setting, one might wonder whether much of the performance gap is due to the poor quality of automatically generated bounding-boxes. In contrast, still image detectors are typically trained from manually drawn bounding-boxes.
%
The second factor we consider is the {\em appearance diversity} of the training samples within a domain (sec.~\ref{sub:AD}). Video differs from still images in that frames are temporally correlated. Frames close in time often contain near identical samples of the same object, whereas in still image datasets such repetition happens rarely. This is an intrinsic difference in the medium, and leads to this often overlooked factor. 
%
The next factor is {\em image quality} (sec.~\ref{sub:IQ}). Video is typically more blurry than images, e.g. large motion blur smoothens video frames in a particular direction. Differences in compression schemes or color contrast between the two domains  also affect image quality. In practice, we measure Gaussian blur and motion blur as representative for this factor.
%
The fourth factor is the {\em distribution over aspects}, i.e. the type of object samples in the training sets (sec.~\ref{sec:aspects}). As the space of possible samples for an object class is very large, each dataset covers it only partially~\cite{torralba2011cvpr}, with its own specific bias. For example, horses jumping over hurdles might appear in one dataset but not in another. Hence, an important factor is the differences in the aspect distributions between the two domains.
%
Finally, we consider {\em object size and camera framing} issues (sec.~\ref{sec:other_factors}).
Photographers and videographers might follow different approaches when capturing an object, e.g. in images the objects tend to be fully in focus, while videos might have objects coming in and out of the frame. Also the distance at which objects are captured might be different. Hence, we considered the distribution of object size, aspect-ratio, and truncation by the image frame as a last factor.

\begin{figure}
\centering
\includegraphics[width=\linewidth]{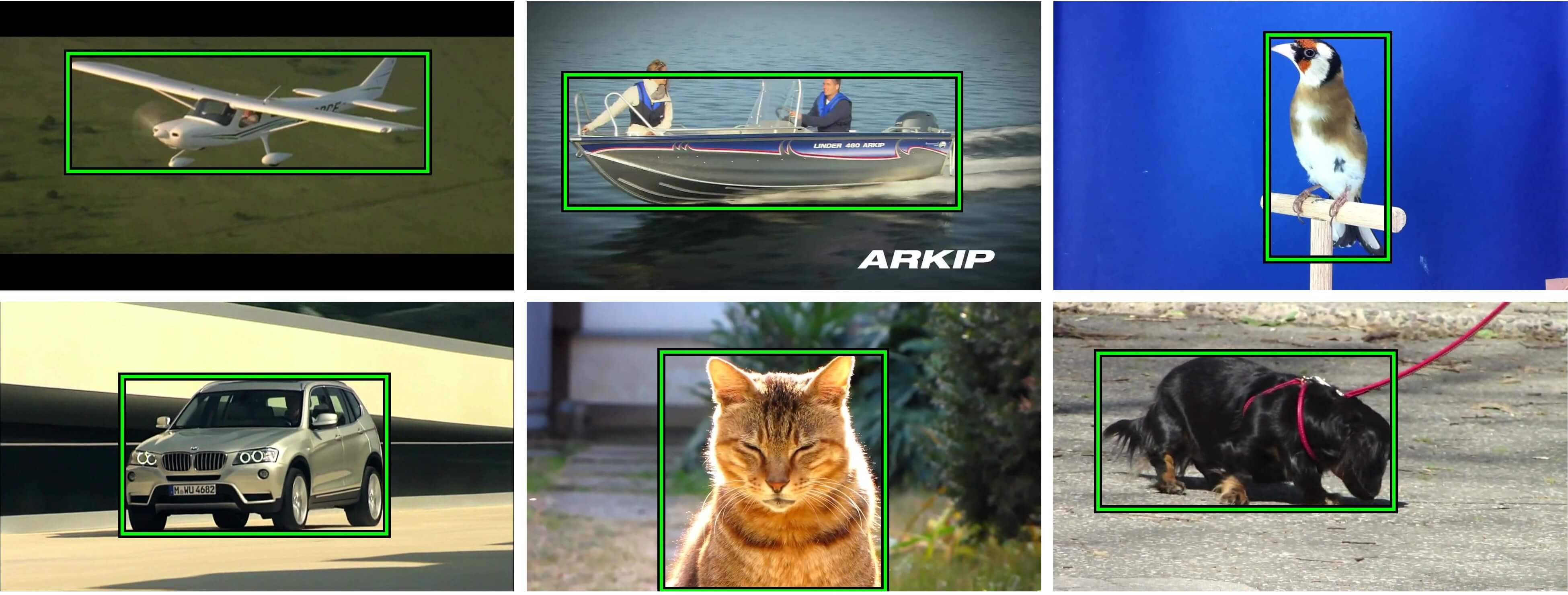}
\vspace*{-6mm}
\caption{\textit{Example video frames with ground-truth bounding-box annotations.}} 
\vspace{-7mm}
\label{figure:dataset}
\end{figure}

We proceed by examining and evaluating each domain shift factor in turn, following the same structure:
(1) we introduce a metric to quantify the factor in each domain;
(2) we 
modify the training set of each domain so that they are more similar 
in terms of this metric, effectively cancelling out the factor;
(3) we examine the impact of this equalization on the performance of the object detector.
As we found no difference in object size and camera framing between the two datasets (sec.~\ref{sec:other_factors}), we carry out this procedure for the first four factors.

We consider the {\em performance gap}, i.e. the difference in performance of a detector between training on video frames and on still images (on a fixed test set consisting of still images).
We examine the evolution of the performance gap as the training sets get progressively equalized by the procedure above. We also repeat the study in the reverse direction, i.e. where the test set is fixed to video frames.
The results show that all factors affect detection performance and that cancelling them out helps bridging the performance gap.
We perform experiments on two popular object detection models~\cite{felzenszwalb10pami,girshick14cvpr}. 
While these are very different, our results hold for both, suggesting that our findings apply to object detection in general.
Moreover, the results follow the same trends on both dataset pairs we considered, showing that the domain shift factors we examine are relevant in general,  and the effects we observe are not specific to a particular dataset.

\begin{table}[t]
\centering
\vspace{-2mm}
\caption{Number of object samples in the training and test sets for image (VOC) and video (YTO) domains.}  
\vspace{-3mm}
\label{tab:all_trainingtest}
{\footnotesize
\begin{tabular}{ |c||c|c|c||c|c|}
  \hline
  \multirow{3}{*}{Classname}  & \multicolumn{3}{|c||}{Training} &\multicolumn{2}{|c|}{Test}\\
  \cline{2-6}
  &  \multicolumn{3}{|c||}{Number of object samples} &
  \multicolumn{2}{|c|}{Number of object samples}\\
  \cline{2-6}
    & VOC & YTO  & Equalized  &VOC & YTO \\
  \cline{2-6}  
\hline
aeroplane & 306 & 415 & 306 & 285 & 180\\ 
\hline
bird & 486 & 359 & 359 & 459 & 162\\ 
\hline
boat & 290 & 357 & 290 & 263 & 233\\ 
\hline
car & 1250 & 915 & 915 & 1201 & 605\\ 
\hline
cat & 376 & 326 & 326 & 358 & 165\\ 
\hline
cow & 259 & 321 & 259 & 244 & 315\\ 
\hline
dog & 510 & 454 & 454 & 489 & 173\\ 
\hline
horse & 362 & 427 & 362 & 348 & 463\\ 
\hline
motorbike & 339 & 360 & 339 & 325 & 213\\ 
\hline
train & 297 & 372 & 297 & 282 & 158\\ 
\hline \hline
total & 4475 & 4306 & 3907 & 4254 & 2667\\ 
\hline
\end{tabular}
\vspace{-4mm}
}
\end{table}

\vspace{-2mm}
\section{Datasets and protocol}
\label{sec:dataset}

In this section and the next we focus on the first dataset pair (i.e. PASCAL VOC 2007 and YouTube-Objects). Results on the second pair (ILSVRC 2015)  are reported in sec.~\ref{sec:ILSCVRC}.

For still images, we use PASCAL VOC 2007 \cite{pascal07:thomas}, one of the most widely used datasets for
object detection. For video frames we employ YouTube-Objects~\cite{prest12cvpr}, which is one of the largest available video datasets with bounding-box annotations on multiple classes. It has $10$ classes from PASCAL VOC 2007, which enables studying image-video domain differences.
We train two modern object detectors~\cite{girshick14cvpr,DPMrelease5}
with annotated instances either from still images or from video frames
and test them on both domains. In this fashion, we can observe how
the performance of a detector depends on the domain it
is trained from.

\vspace{-2mm}
\subsection{Datasets}
\label{sub:da}

\begin{figure*}[!htb]
\ifx
\centerline{%
\begin{tabular}{c@{}c@{}c@{}c@{}c@{}}
\includegraphics[width=0.50\textwidth]{newfigs/DPM_VOC.pdf}&
\textit{ } &
\includegraphics[width=0.50\textwidth]{newfigs/DPM_VID.pdf} \\
\small{(a)} & \textit{ } & \small{(b)} \\
\vspace{-6mm}
\includegraphics[width=0.50\textwidth]{newfigs/R-CNN_VOC.pdf}&
\textit{ } &
\includegraphics[width=0.50\textwidth]{newfigs/R-CNN_VID.pdf} \\
\small{(c)} & \textit{ } & \small{(d)} \\
\end{tabular}}
\fi
\centerline{%
\begin{tabular}{c@{}c@{}c@{}c@{}c@{}}
\includegraphics[width=0.50\textwidth]{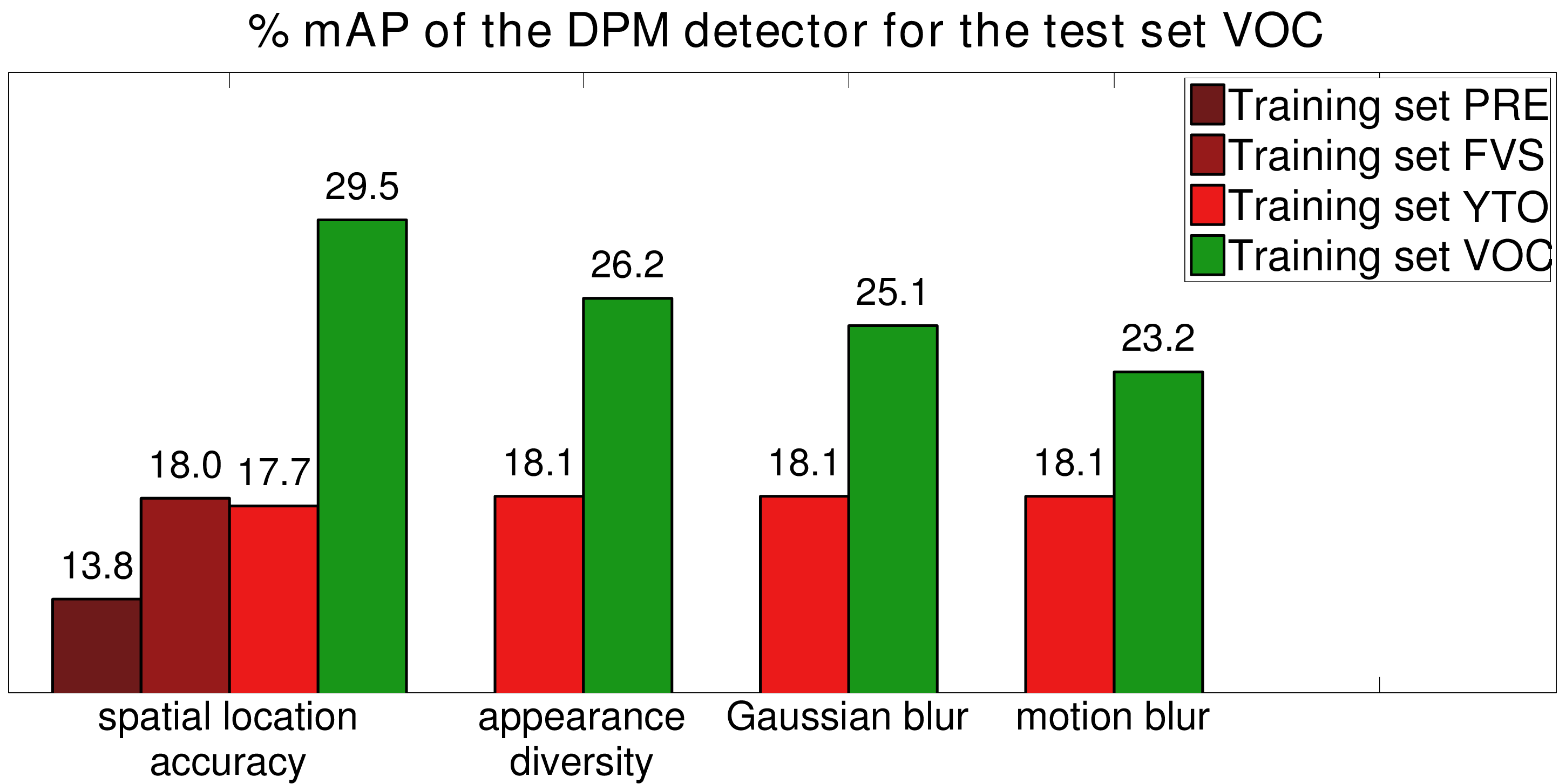}&
\textit{ } &
\includegraphics[width=0.50\textwidth]{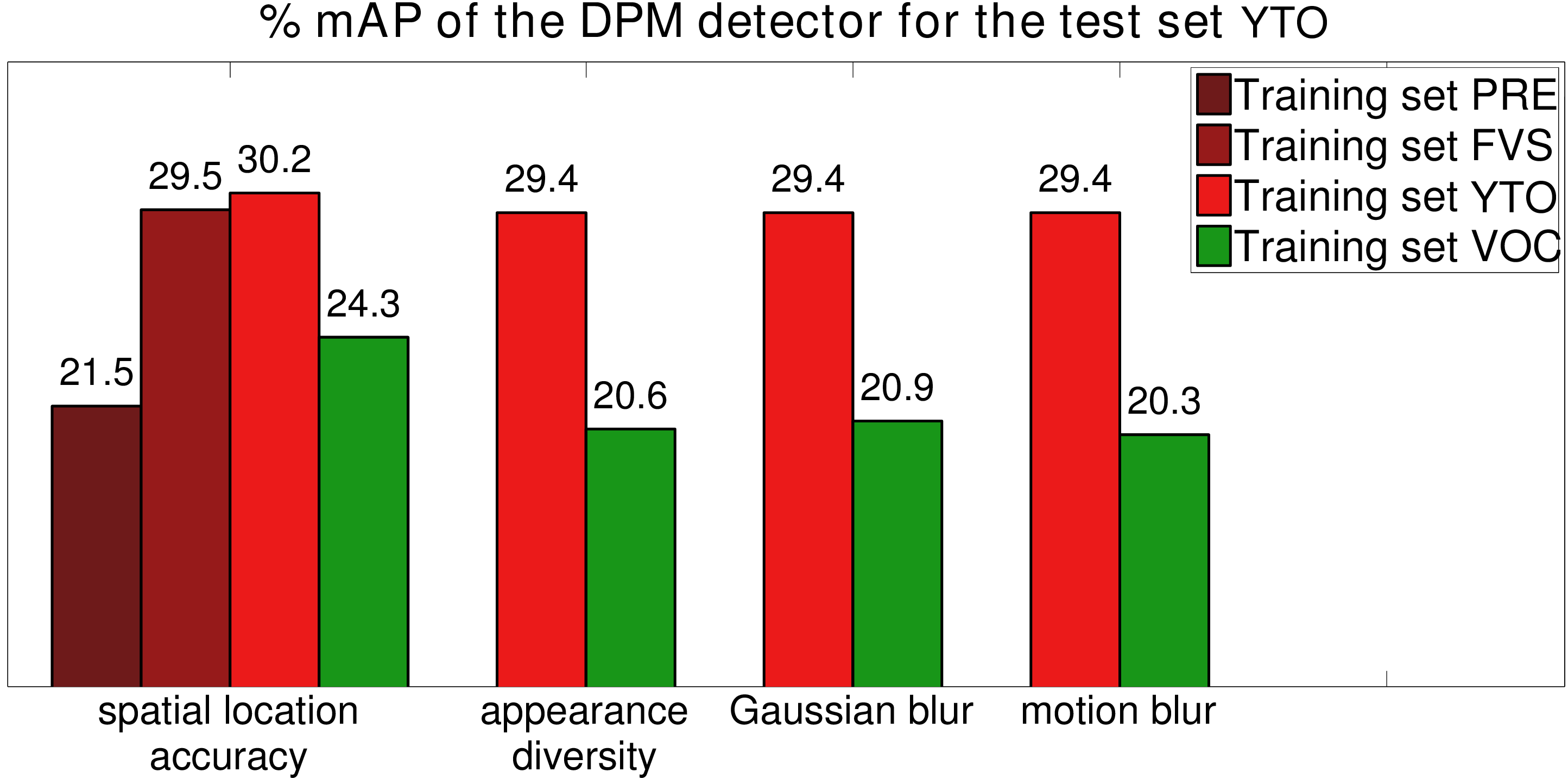} \\
\vspace{-2mm}
\includegraphics[width=0.50\textwidth]{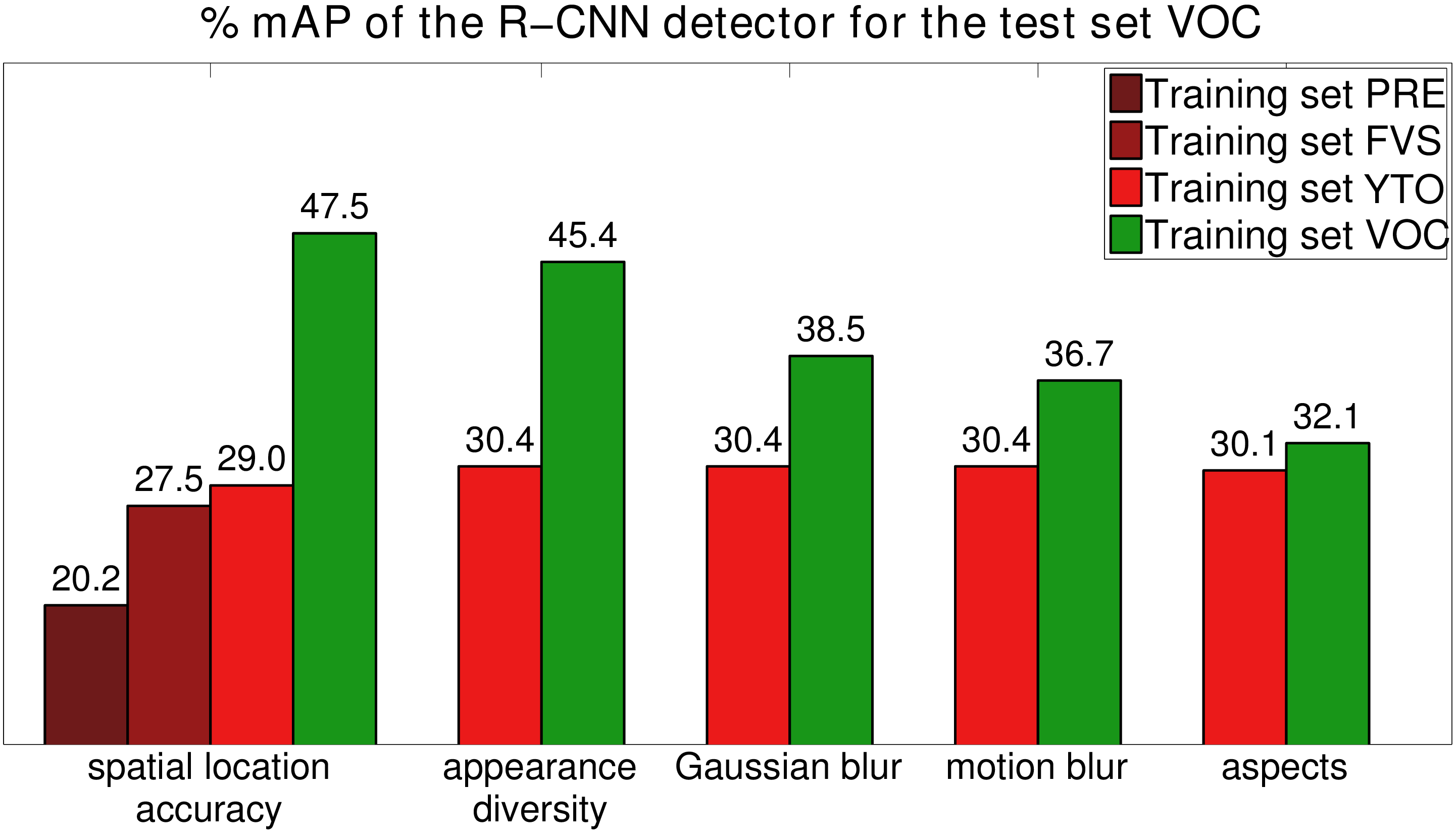}&
\textit{ } &
\includegraphics[width=0.50\textwidth]{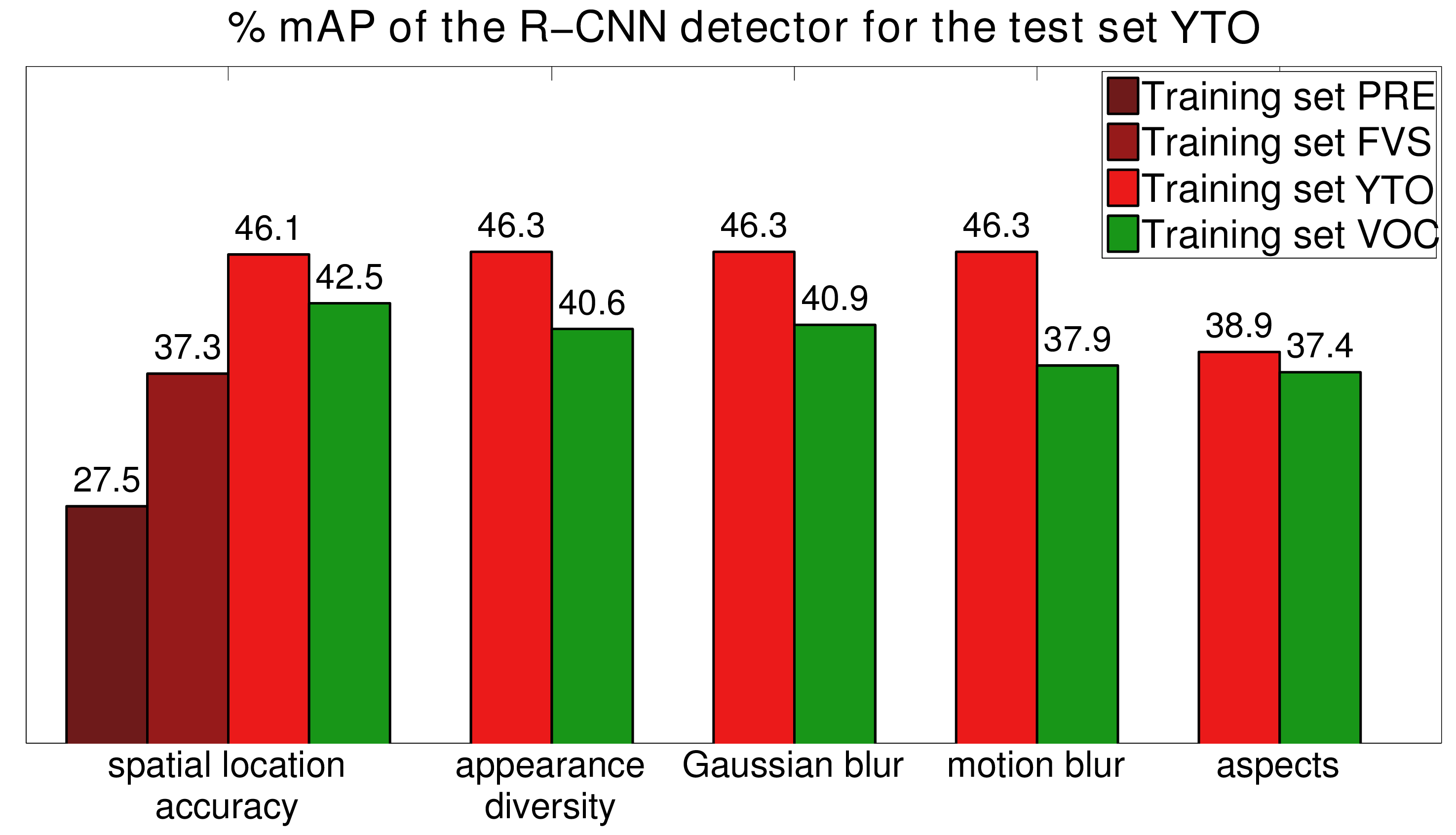} \\
\end{tabular}}
\vspace{-2mm}
\caption{\textit{Impact of the domain shift factors when training on VOC and
  YTO for two detectors DPM (top row (a) and (b)) and R-CNN (bottom row (c) and (d)) and for
  two test sets VOC (left column) and YTO (right column). }
}
\vspace{-4mm}
\label{figure:map}
\end{figure*}

\paragraph*{Still images (VOC)} \label{VOC}
Out of the $20$ classes in PASCAL VOC 2007, we use the $10$ which have moving objects, in order to have the same ones as in YouTube-Objects.
Each object instance of these classes is annotated with a bounding-box in both training and test sets.
Tab.~\ref{tab:all_trainingtest} shows dataset statistics.

\vspace{-1mm}
\paragraph*{Video frames (YTO)} \label{VID}
The YouTube-Objects dataset~\cite{prest12cvpr} contains videos collected from
YouTube for $10$ classes of moving objects. While it consists of $155$ videos over $720,152$ frames, only $1,258$ of them are annotated with a bounding-box around an object instance ($3\times$ fewer than in VOC).
Instead, we would like to have a comparable number of annotations in both datasets. This would exclude differences in performance due to differences in the size of the training sets.
Hence, we annotate many additional bounding-boxes on frames from YouTube-Objects. We first split the videos into disjoint training and test sets. In order to avoid any bias between training and test set, frames from the same video belong only to one set.
Then, for both sets, we uniformly sample a constant number of frames in each shot, so that the total number of YTO training samples is roughly equal to the number of VOC training samples.
For the training set, we annotate one object instance per frame. For the test
set, we annotate all instances. The total number of annotated samples is
$6,973$ (obtained from $6,087$ frames). Fig~\ref{figure:dataset} shows some
annotated frames. The additional annotations are available on-line at \href{http://calvin.inf.ed.ac.uk/datasets/youtube-objects-dataset/}{http://calvin.inf.ed.ac.uk/datasets/youtube-objects-dataset/}.

\paragraph*{Equalizing the number of samples per class} 
For each class, we equalize the number of training samples {\em exactly}, by randomly sub-sampling the larger of the two training sets. The final number of equalized training samples is
$3,907$ in total over the $10$ classes (see column `equalized' in tab.~\ref{tab:all_trainingtest}).
Only these equalized training sets will be used in the remainder of the paper.
We refer to them to as trainVOC and trainYTO for still images and video frames, respectively.  

\vspace{-3mm}
\subsection{Protocol} 
\label{Protocol}
Recall that we want to train object detectors either from
still images or from video frames and then test them on both domains.    
Each training set contains samples from one domain only. For a class,
the positive training set contains annotated samples of this class,
while the negative set contains images of all other classes.
When testing on still images, we use the complete PASCAL VOC 2007 test set (tab.~\ref{tab:all_trainingtest}; this includes also images without instances of our 10 classes).
When testing on video, we use a test set {of $1,781$ images with $2,667$ objects instances in total (tab.~\ref{tab:all_trainingtest}).
We refer to them as testVOC and testYTO, respectively. 

We measure performance using the PASCAL VOC protocol. A detection is correct if its intersection-over-union overlap with a ground-truth bounding-box is $>0.5$ \cite{pascal07:thomas}.
The performance for a class is Average Precision (AP) on the test set, and the overall performance is captured by the mean AP over all classes (mAP).

We experiment using two modern object detectors: the Deformable Part Model
(DPM)~\cite{felzenszwalb10pami, DPMrelease5} and the Regions with
Convolutional Neural Networks (R-CNN) \cite{girshick14cvpr}.
DPM models an object class by a mixture of components, each composed of a root HOG template~\cite{Dalal05:thomas} and a collection of part templates arranged in a deformable
configuration. This detector was the state-of-the-art reference for several years, until the arrival of CNN-based models.

R-CNN~\cite{girshick14cvpr} is the current leading object detector. Candidate regions
are obtained by selective search \cite{Uijlings13} and described with
convolutional neural networks features (CNNs) extracted with Caffe \cite{jia2014caffe, jia2014caffecode}.
A linear SVM is then trained to separate positive and negative training regions (with hard negative mining to handle the large number of negative regions \cite{Dalal05:thomas,girshick14cvpr,DPMrelease5}).
In this paper, we use as features the $7^{th}$ layer of the CNN model trained on the ILSVRC12 classification challenge \cite{Krizhevsky2012nips}, as provided by~\cite{girshick14cvprcode}. We do not fine-tune the CNN for object detection, so that the features are not biased to a particular dataset. This enables to measure domain shift factors more cleanly.

\vspace{-2mm}
\section{Domain shift factors}
\label{sec:factors}

In this section we analyse the difference between VOC and YTO according to four
factors: spatial location accuracy, appearance diversity, image
quality and aspect distribution.
We examine each factor by following the same procedure:
({\em 1: measurement}) We introduce a metric to quantify the factor in each domain.
({\em 2: equalization}) We present a way to make the training sets of the two domains more similar in terms of this metric.
({\em 3: impact}) We compare the performance of object detectors trained from each domain before and after the equalization step. This enables to measure if, and by how much, equalization reduces the performance gap due to training on different domains.

As we apply the procedure above to each factor in sequence,
we observe the evolution of the performance gap as the two domains are gradually equalized.
As we have two test sets (one per domain) we monitor the evolution of two performance gaps in parallel. 


\vspace{-2mm}
\subsection{Spatial location accuracy}
\label{sub:SLA}

There are several methods to automatically segments objects from the background in video frames by exploiting spatio-temporal continuity \cite{brox10eccv,lee11iccv,Papazoglou_2013_ICCV,prest12cvpr}.
We evaluate two methods:
(PRE) the method of \cite{prest12cvpr}, which extends the motion segmentation algorithm~\cite{brox10eccv} to joint co-localization over all videos of an object class;
and
(FVS) the fast video segmentation method of \cite{Papazoglou_2013_ICCV}, which operates on individual videos.
Both methods automatically generate bounding-boxes for all video frames.
We sample as many bounding-boxes as there are in the trainYTO and trainVOC sets by following the approach in \cite{prest12cvpr}.
In the first step we quantify the quality of each bounding-box, based on its objectness probability~\cite{alexe12pami} and the amount of contact with the image border (boxes with high contact typically contain background).
In the second step we randomly sample bounding-boxes according to their quality (treating the quality values for
all samples as a multinomial distribution).
In this way, we obtain the PRE and FVS training sets.

In this section, we use the trainVOC set for still images. For
video frames, we use the PRE and FVS training sets and we measure
their accuracy with respect to the ground-truth annotations
(sec.~\ref{sub:SLA}: \hyperref[sub:SLA_M]{Measurement} 
). We also use the trainYTO set, in order to improve video training data to match the perfect spatial support of still images (sec.~\ref{sub:SLA}:\hyperref[sub:SLA_M]{Equalization}).
Finally, we train object detectors from each training set and test them on testVOC and testYTO. In this way, we can quantify the impact of the different levels of spatial location accuracy on performance (sec.~\ref{sub:SLA}: \hyperref[sub:SLA_M]{Impact}).  

\ifx
\begin{figure}[t]
\centerline{%
\begin{tabular}{c@{}c@{}c@{}c@{}c@{}}
\includegraphics[width=0.50\linewidth]{figures/54.jpg}&
\textit{ } &
\includegraphics[width=0.50\linewidth]{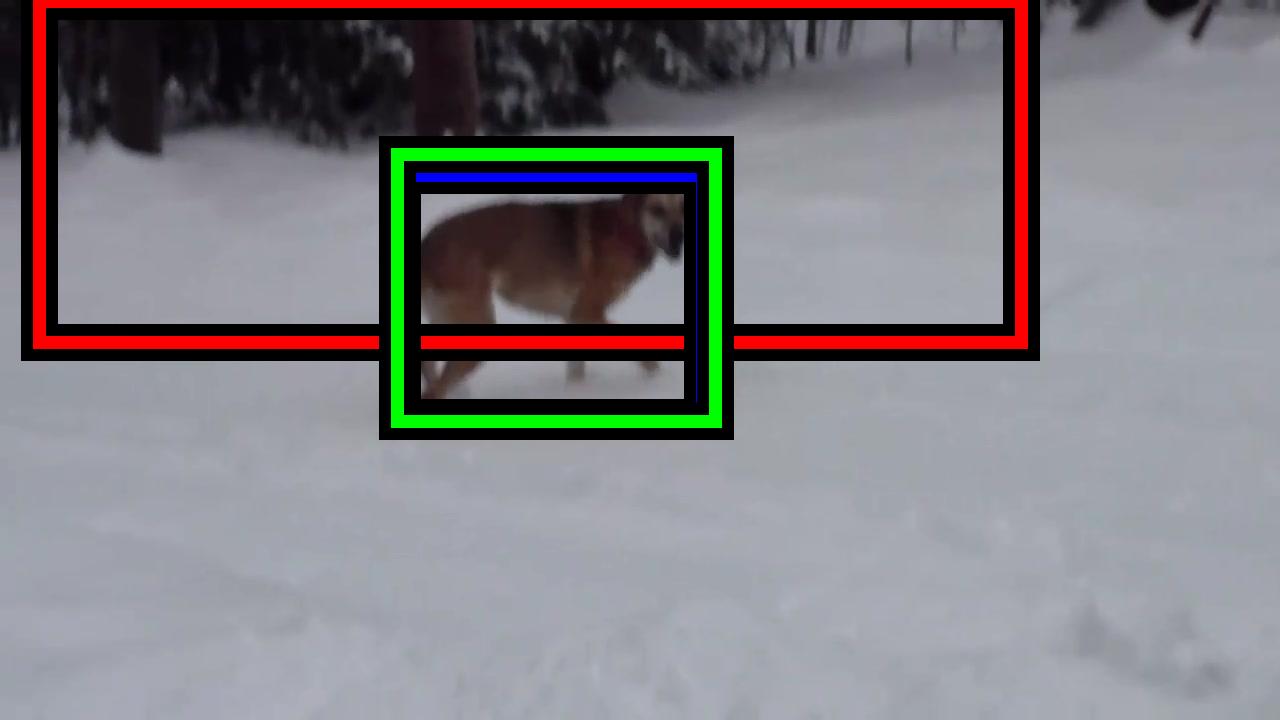} \\
\includegraphics[width=0.50\linewidth]{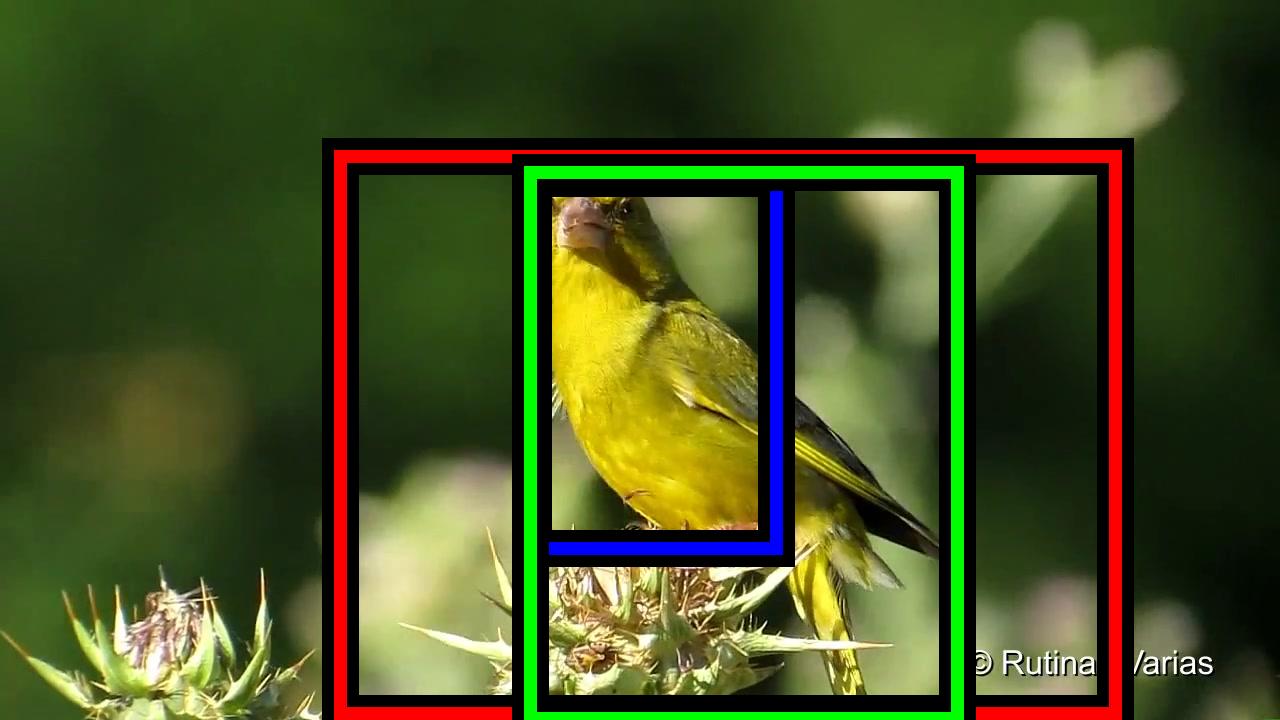}&
\textit{ } &
\includegraphics[width=0.50\linewidth]{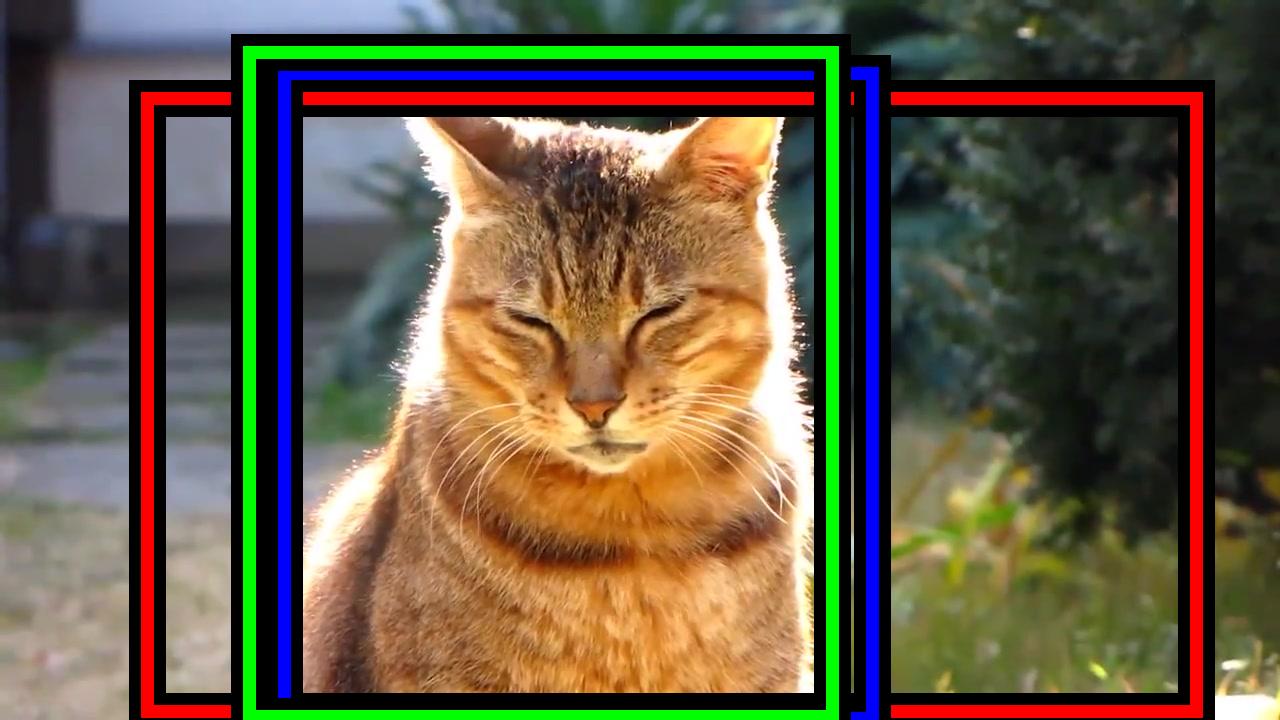} \\
\end{tabular}}
\vspace*{-2mm}
\caption{\textit{Example bounding-boxes produced by PRE~\cite{prest12cvpr} (red), FVS~\cite{Papazoglou_2013_ICCV} (blue), and ground-truth annotations (green).}}
\vspace{-4mm}
\label{figure:sla}
\end{figure} 
\fi

\begin{figure}[t]
\centerline{%
\begin{tabular}{c@{}c@{}c@{}c@{}c@{}}
\includegraphics[width=0.32\linewidth]{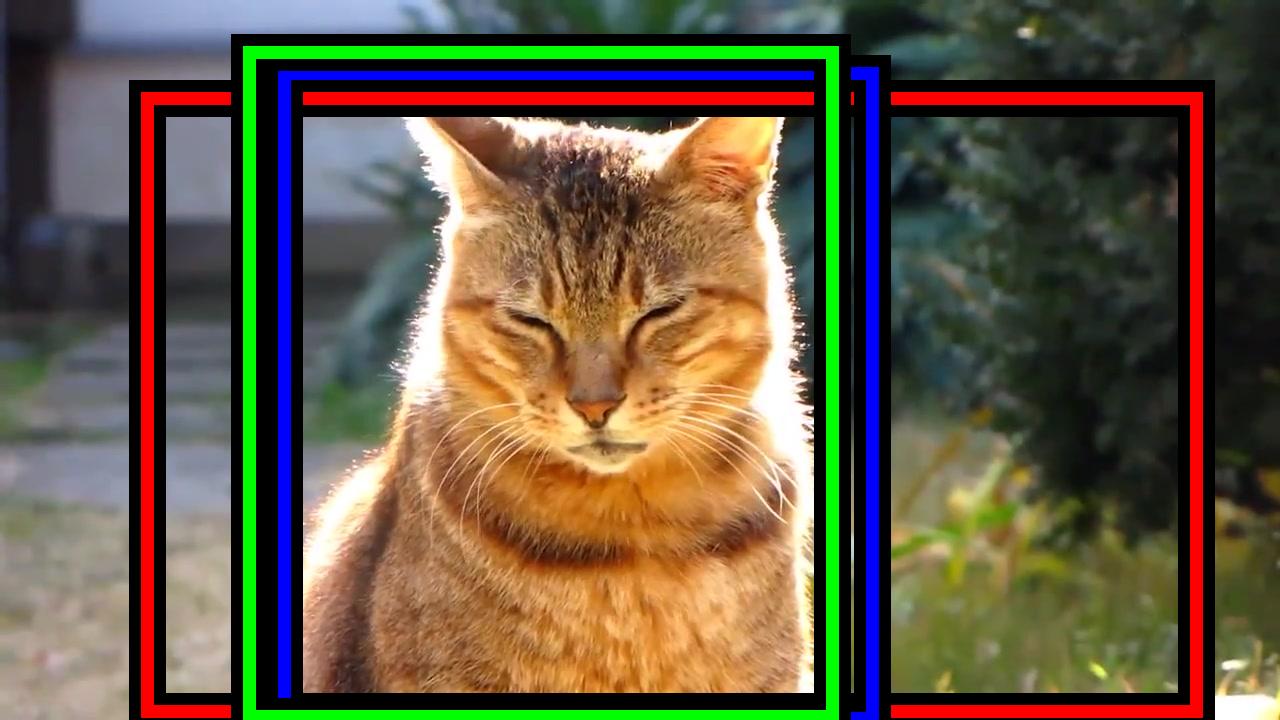}&
\textit{ } &
\includegraphics[width=0.32\linewidth]{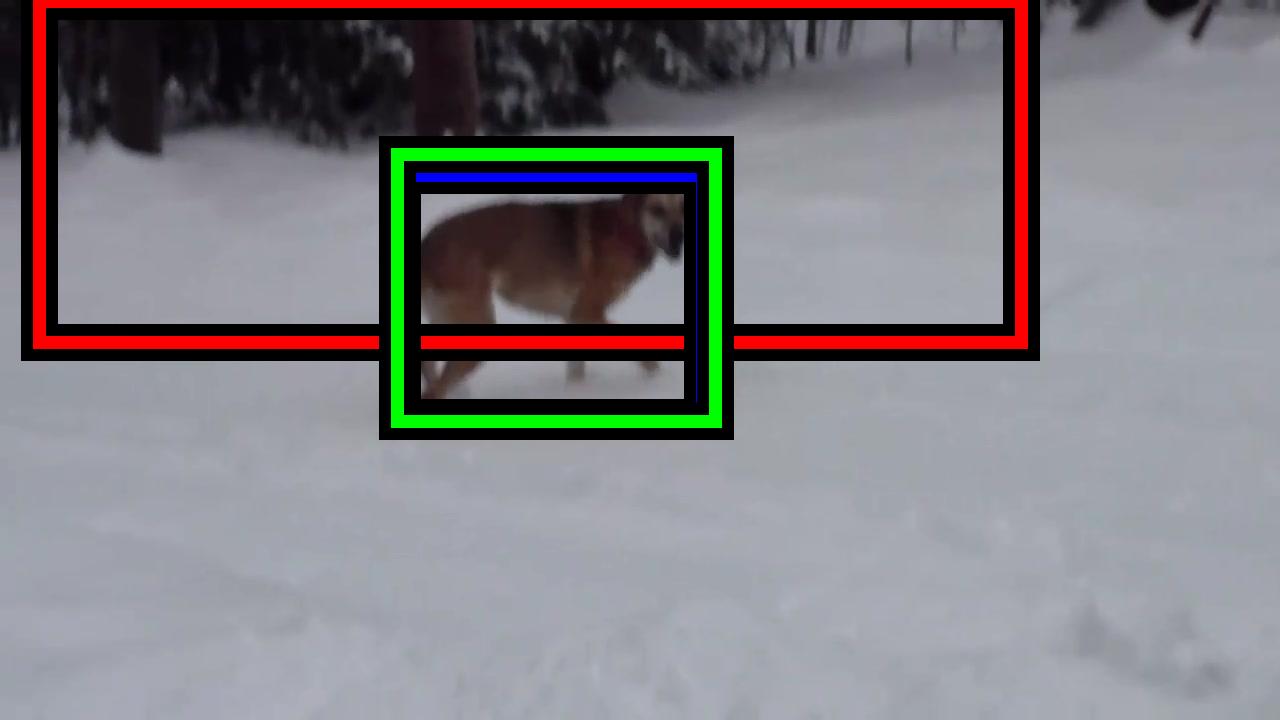} &
\textit{ } &
\includegraphics[width=0.32\linewidth]{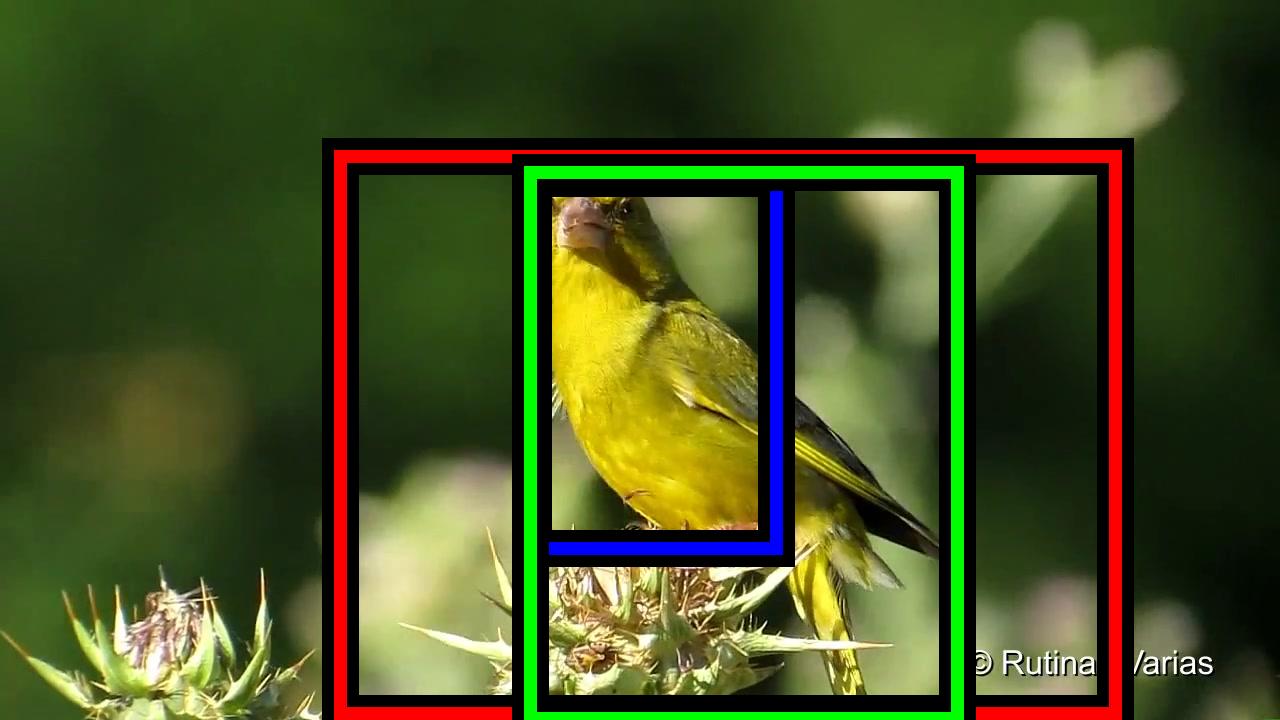}\\
\end{tabular}}
\vspace*{-2mm}
\caption{\textit{Example bounding-boxes produced by PRE~\cite{prest12cvpr} (red), FVS~\cite{Papazoglou_2013_ICCV} (blue), and ground-truth annotations (green).}}
\vspace{-4mm}
\label{figure:sla}
\end{figure} 

\paragraph*{Measurement} 
\phantomsection
\addcontentsline{toc}{section}{Measurement}
\label{sub:SLA_M}

\ifx
\begin{figure}[t]
\begin{center}
\includegraphics[width=\linewidth]{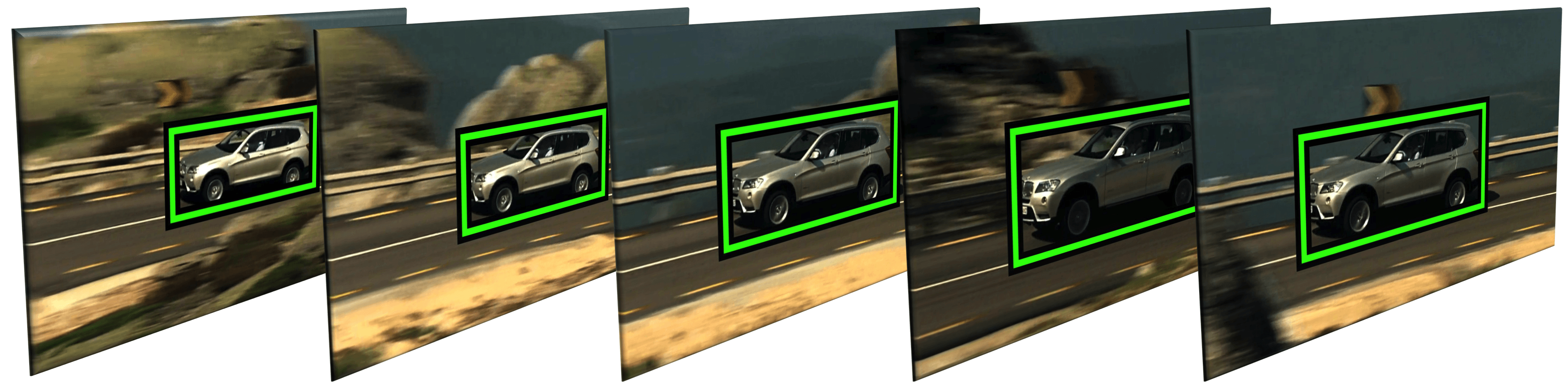}
\end{center}
\caption{\textit{YTO dataset: Frames in the same shot that contain near identical samples of an object.} }
\label{figure:UnEx}
\vspace{-4mm}
\end{figure}

\begin{figure}[t]
\begin{center}
\includegraphics[width=\linewidth]{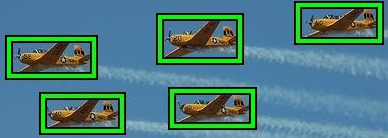}
\caption{\textit{VOC dataset: Example of near identical samples in the same image.} }
\label{figure:ApDi1VOC}
\end{center}
\vspace{-4mm}
\end{figure}
\fi

We measure the accuracy of bounding-boxes by CorLoc: the percentage of
bounding-boxes that satisfy the PASCAL VOC criterion \cite{Everingham10}
(IoU $>50\%$). Bounding-boxes delivered by the PRE method have $24.0\%$ CorLoc, while FVS brings $54.3\%$ CorLoc.
This shows that FVS can automatically produce good bounding-boxes in about half the frames, which is considerably better than the older method~\cite{prest12cvpr} (fig.~\ref{figure:sla}).
However, this is worse than having all frames correctly annotated (as it is the case with manual ground-truth).

\paragraph*{Equalization} 
\phantomsection
\addcontentsline{toc}{section}{Equalization}
\label{sub:SLA_E}

The equalization step enhances the quality of the bounding-boxes in video frames, gradually moving from the worst to perfect annotations.
We match the perfect location accuracy of the still image trainVOC set by using ground-truth bounding-boxes for the video frames (trainYTO set).

\paragraph*{Impact} 
\phantomsection
\addcontentsline{toc}{section}{Impact}
\label{SLA_I}

For video frames we train object detectors for each of the three levels of spatial support: starting
with poor automatic annotations (PRE), then moving to better ones (FVS), and finally using ground-truth bounding-boxes (trainYTO). For still images we train detectors from the trainVOC set (ground-truth bounding-boxes). We test on the testVOC and testYTO sets. Fig.~\ref{figure:map} reports the performance for both detectors (DPM, R-CNN) and test sets (testVOC, testYTO).

When testing on still images (testVOC), the mAP of training from video continuously improves when using more and more accurate spatial support (fig.~\ref{figure:map}a,c). 
However, the performance of training on trainVOC is still considerably superior even to training on trainYTO with perfect ground-truth annotation.
These results shows that the imperfect spatial location accuracy of training samples produced by automatic video segmentation methods can only explain part of the gap.
This is surprising, as we expected that using perfect annotations would close the gap much more.
Quantitatively, for DPM the gap goes from $15.7\%$ to $11.8\%$ when going from training on the weakest 
automatic segmentation (PRE) to ground-truth bounding-boxes (trainYTO).
The result is analog for R-CNN, with the gap going from $27.3\%$ when using PRE, to $18.5\%$ when using trainYTO.
These results implies that we cannot get detectors learned from video to perform very well on still images even with great future progress on video segmentation, and in fact not even by manually annotating frames.
Moreover, this also suggests there are other significant causes that produce the leftover gap.

Testing on video (testYTO) reveals a similar trend: more accurate spatial support on video frames leads to better performance (fig.~\ref{figure:map}b,d). 
Interestingly, training from video here performs better than training from still images (when both are ground-truth annotated). This shows we are confronted with a real domain adaptation problem, where it is always better to train on the test domain. Again results hold for both detectors, but the `reverse gap' left after equalizing spatial location accuracy is smaller than on testVOC: 5.9\% mAP for DPM and 3.6\% for R-CNN.

\begin{figure}[t]
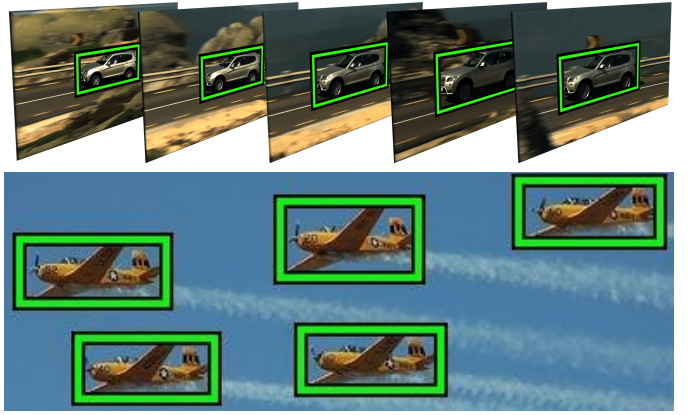

\centerline{%
\begin{tabular}{c@{}c@{}c@{}c@{}c@{}}
\includegraphics[width=\linewidth]{ue3.png} \\
\includegraphics[width=\linewidth]{ApDi1VOCAll4.png} \\
\end{tabular}}
\vspace*{-2mm}
\caption{\textit{(top row) YTO dataset: Frames in the same shot that contain near identical samples of an object. (bottom row) VOC dataset: Example of near identical samples in the same image.}}
\vspace{-4mm}
\label{figure:NearIdentical}
\end{figure}

\vspace{-2mm}
\subsection{Appearance diversity}
\label{sub:AD}

Video is intrinsically different from still images in that frames are temporally correlated. Frames that are close in time often contain near identical samples of the same object 
(top row of fig.~\ref{figure:NearIdentical}. 
In still images such repetition happens rarely and typically samples that look very similar co-occur in the same image 
(bottom row of fig.~\ref{figure:NearIdentical}).
We first measure the appearance diversity of training sets 
(sec.~\ref{sub:AD}: \hyperref[sub:SLA_M]{Measurement}). 
Then we modify them to equalize their appearance diversity
(sec.~\ref{sub:AD}: \hyperref[sub:SLA_M]{Equalization}).
Finally, we observe the impact of this equalization on the
performances of object detectors 
(sec.~\ref{sub:AD}: \hyperref[sub:SLA_M]{Impact}).
In the spirit of our progressive equalization mission, here we use the trainYTO and trainVOC sets, which have ground-truth annotations. In this way, we focus on differences due to appearance diversity alone and not due to spatial support. 

\vspace{-1mm}
\paragraph*{Measurement} 
\phantomsection
\addcontentsline{toc}{section}{Measurement}
\label{AD_M}

To measure appearance diversity within a training set, we manually group near-identical training samples, i.e. samples of identical objects in very similar viewing conditions (e.g. viewpoint and degrees of occlusion, fig.~\ref{figure:NearIdentical}}). 
This results in a set of groups, each containing near-identical samples (fig~\ref{figure:UnExClusters}).
We quantify appearance diversity by the number of groups, i.e. the number of {\em unique samples} in the training set.

As shown in tab.~\ref{tab:ue}, trainYTO has only half the number of unique samples than trainVOC, despite them having exactly the same total number of samples (tab.~\ref{tab:all_trainingtest}).
This shows that half of the video samples ($51\%$) are repeated, while almost all ($97\%$) still images samples are unique. This reveals a considerable difference in appearance diversity between the two domains.

\vspace{-1mm}
\paragraph*{Equalization} 
\phantomsection
\addcontentsline{toc}{section}{Equalization}
\label{AD_E}

We equalize appearance diversity by resampling each training set so that:
(1) it contains only unique samples;
and
(2) the size of the training sets is the same in the two domains.
We achieve the first goal by randomly picking one sample per group,
and the second by randomly subsampling the larger of the two training sets (i.e. VOC).
This procedure is applied for each class separately.
This leads to the new training sets `trainVOC Unique Samples' and `trainYTO
Unique Samples', each containing $2,201$ unique samples (tab.~\ref{tab:ue}, column `Equalized Unique Samples').

\vspace{-1mm}
\paragraph*{Impact} 
\phantomsection
\addcontentsline{toc}{section}{Impact}
\label{AD_I}

We train object detectors from the equalized unique sample sets only. Fig.~\ref{figure:map} reports results for both detection models and test sets.
%
When testing on VOC, the mAP of training from still images decreases significantly when going from using all training samples (trainVOC) to trainVOC Unique Samples, as about half of the unique training samples are removed. Instead, the mAP of training from video remains almost constant, as only duplicate samples are removed.
Testing on YTO produces similar effects, with the unique sample equalization procedure leaving the performance of training from YTO almost unchanged, but significantly reducing that of training from VOC.
These results reveal that indeed near identical samples do not bring any extra information, and only artificially inflate the apparent size of a training set.
Hence, these findings suggest that one should pool training samples out of a large set of diverse videos, sampling very few frames from each shot.

Equalizing appearance diversity reduces the performance gap when testing on VOC down to $8.1\%$ mAP for DPM and $15.0\%$ mAP for R-CNN. Notably, this bridges the gap for both detectors by about the same amount ($3.5\%-3.7\%$).
When testing on YTO the equalization has the opposite effect and increases the gap by about 3\% to $8.8\%$ mAP for DPM and $5.7\%$ for R-CNN. This makes sense, as the process handicaps trainVOC down to the level of diversity of trainYTO, without harming trainYTO.

\begin{figure}[t]
\begin{center}
\includegraphics[width=\linewidth]{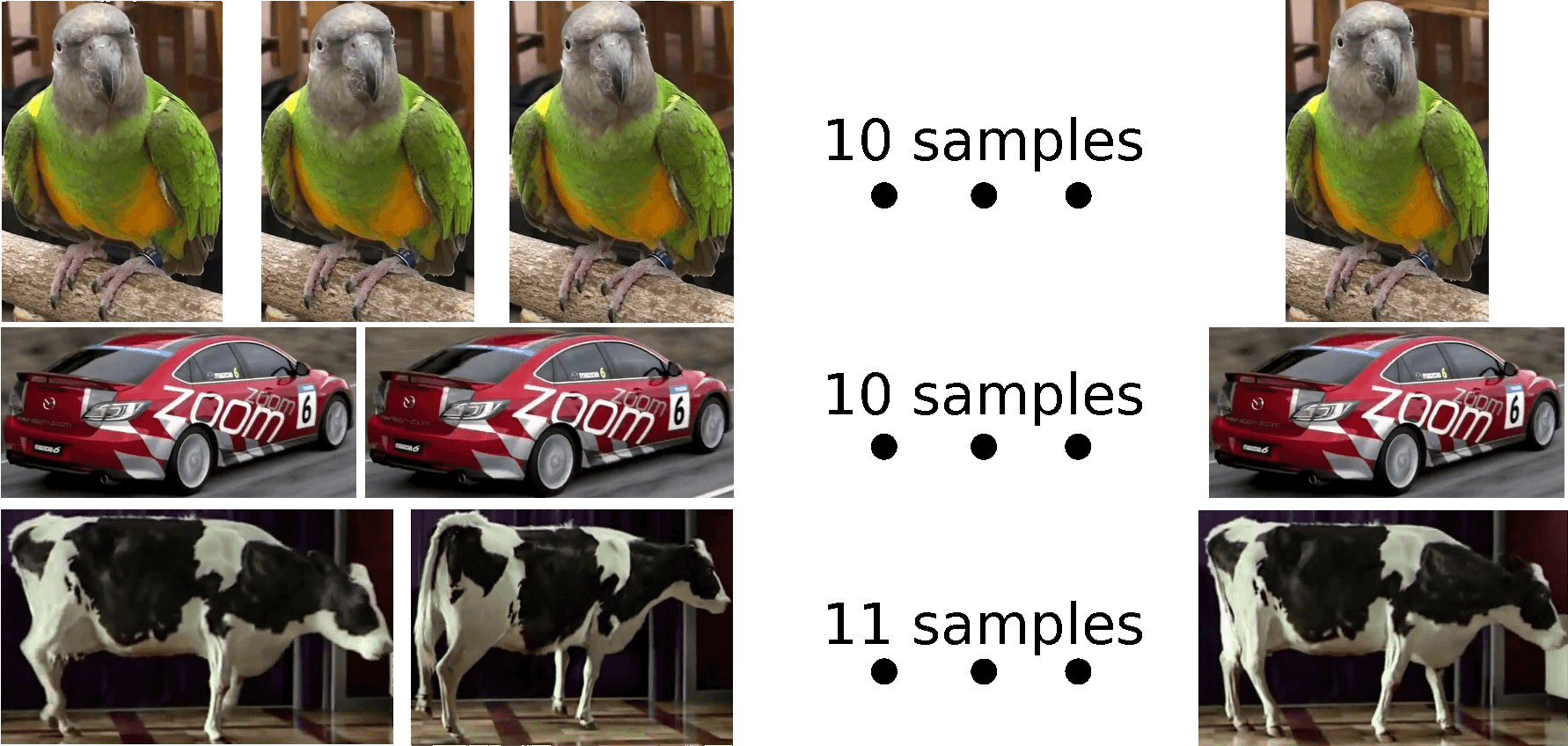}
\end{center}
\vspace*{-2mm}
\caption{\textit{Three example groups of near-identical samples in trainYTO. We display a subset of the frames for each group.}}
\vspace{-4mm}
\label{figure:UnExClusters}
\end{figure}

\vspace{-2mm}
\subsection{Image quality}
\label{sub:IQ}

We examine the image quality factor while working on the Unique Samples training sets,
which have the same size, accuracy of spatial support, and level of appearance diversity. In this way, all those factors will not cause any performance difference. 

\begin{figure*}[t]
\begin{center}
\includegraphics[ width = \textwidth]{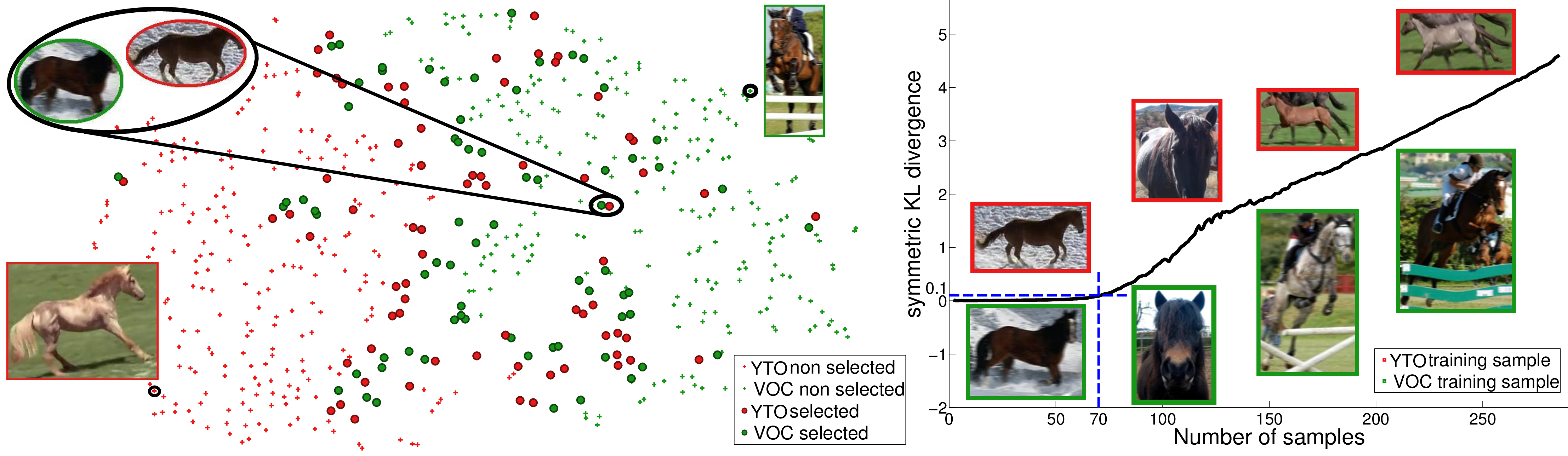}
\vspace*{-2mm}
\caption{\textit{
(Left) 2D visualization of the trainYTO Unique Samples (red) and trainVOC Motion Blurred Unique Samples (green) for the `horse' class in R-CNN feature space. Circles indicate the samples selected by our equalization technique of sec.~\ref{sec:aspects}: Equalization. 
(Right) evolution of $d_{KL}$ as more and more sample pairs are added by our algorithm. The two distributions are very similar at the beginning and start to diverge later. At the $\epsilon=0.1$ threshold, 70 samples from each set are selected (left). This number is driven by $\epsilon$ and changes from class to class.
}}
\vspace{-3mm}
\label{figure:horse_kl}
\end{center}
\vspace{-4mm}
\end{figure*} 

\vspace{-1mm}
\paragraph*{Measurement}
\phantomsection
\addcontentsline{toc}{section}{Measurement}
\label{IQ_M}

We measure the image quality of a training sample by its gradient energy, as in \cite{prest12cvpr}.
This computes the sum of the gradient magnitudes in the HOG cells of an object bounding-box, normalized by its size (computed using the implementation of~\cite{felzenszwalb10pami}).
The gradient energy averaged over all classes is $4.4$ for trainVOC Unique Samples and $3.2$ for trainYTO Unique Samples.
This is because video frames suffer from compression artefacts, motion blur, and low color contrast.

\begin{figure}[t]
\begin{center}
\includegraphics[width=\linewidth]{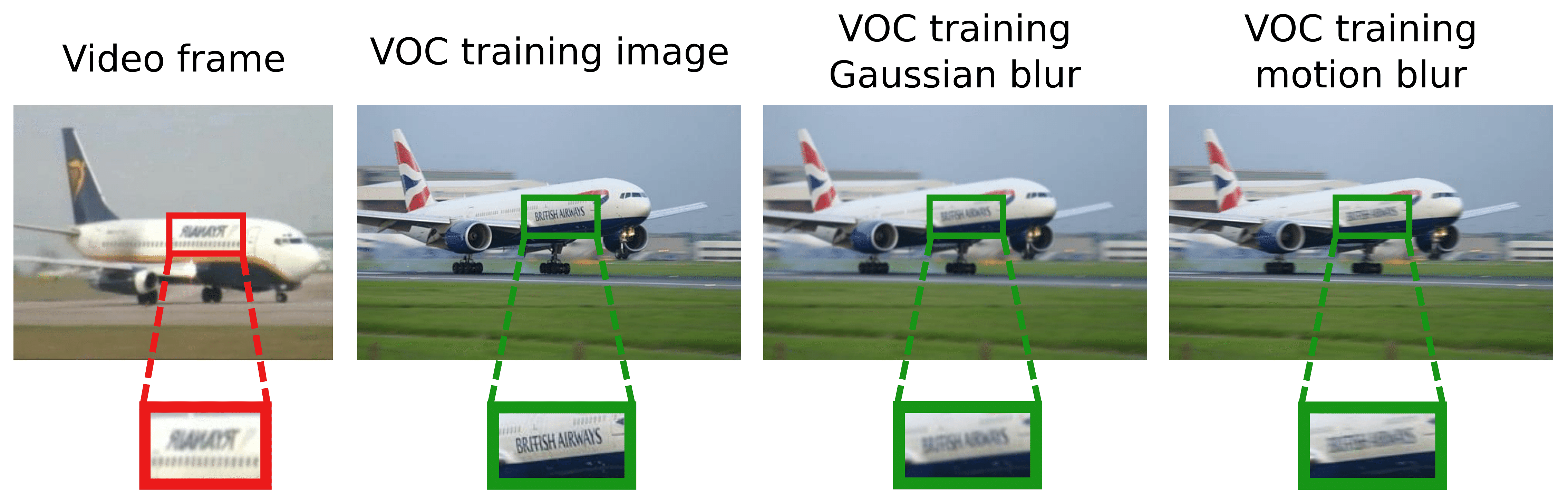}
\vspace*{-4mm}
\caption{\textit{Video frame, VOC training image, Gaussian and motion blurred
  VOC training images.}
}
\vspace{-4mm}
\label{figure:blur}
\end{center}
\end{figure}

\begin{table}[t]
\centering
\vspace{-2mm}
\caption{Appearance diversity equalization. Statistics of the groups: Number of groups, ratio: number of groups / number of ground-truth samples and number of equalized unique samples.}
\vspace{-3mm}
\label{tab:ue}
{\footnotesize
\begin{tabular}{ |c||c|c|c|c|c|}
  \hline  
    \multirow{2}{*}{Classname}  & \multicolumn{2}{|c|}{Number of groups} & \multicolumn{2}{|c|}{Ratio} & Equalized  \\
  \cline{2-5}
  &  YTO & VOC &  YTO & VOC & Unique Samples \\
\hline
aeroplane & 244 & 268 & 0.59 & 0.88 & 244\\ 
\hline
bird & 123 & 452 & 0.34 & 0.93 & 123\\ 
\hline
boat & 138 & 275 & 0.39 & 0.95 & 138\\ 
\hline
car & 310 & 1221 & 0.34 & 0.98 & 310\\ 
\hline
cat & 249 & 376 & 0.76 & 1.00 & 249\\ 
\hline
cow & 90 & 252 & 0.28 & 0.97 &90\\ 
\hline
dog & 295 & 507 & 0.65 & 0.99 & 295\\ 
\hline
horse & 286 & 358 & 0.67 & 0.99 & 286\\ 
\hline
motorbike & 243 & 337 & 0.68 & 0.99 & 243\\ 
\hline
train & 223 & 294 & 0.60 & 0.99 & 223\\ 
\hline \hline
avg & 220 & 434 & 0.51 & 0.97& 220\\ 
\hline
\end{tabular}
\vspace{-3mm}
}
\end{table}

\vspace{-1mm}
\paragraph*{Equalization} 
\phantomsection
\addcontentsline{toc}{section}{Equalization}
\label{IQ_E}

We equalize the gradient energy by blurring the VOC samples, so as to match the energy of the YTO samples. We consider two different ways to blur a sample: Gaussian blur and motion blur.
For Gaussian blur we apply an isotropic Gaussian filter with standard deviation $\sigma$. 
For Motion blur we apply a box filter of length $K$ along the horizontal direction (as most camera motion in YouTube videos is horizontal). 
The motion blurred value $g\left( m, n \right)$ of a pixel
$\left( m, n \right)$ is given by: $g\left(m, n \right) =
\frac{1}{K}  \sum \limits_{ i =0}^{K-1} f \left(m-i,n \right)$.   

We set the parameter of the blur filter ($\sigma$ or $K$) separately for each class, so that the average gradient energy of the blurred VOC samples equals that of the YTO samples. We find the exact parameter values using a bisection search algorithm (as an indication, the average values are $\sigma=1.35$, $K=8.4$).
This procedure leads to the new training sets `trainVOC Gaussian Blurred Unique Samples' and
`trainVOC Motion Blurred Unique Samples'. For uniformity, we also apply the same blur filters to the negative training sets. Fig.~\ref{figure:blur} shows the effect of the blur filters on a VOC training image. 
 
\vspace{-1mm}
\paragraph*{Impact} 
\phantomsection
\addcontentsline{toc}{section}{Impact}
\label{IQ_I}

We train object detectors from either of the two trainVOC blurred Unique Samples sets.
Fig.~\ref{figure:map} report results for both detection models and test sets. Note how results do not change when training from YTO, as this equalization process does not affect video training data.
When testing on VOC, performance drops considerably when using blurred training samples, especially for R-CNN.
On both detection models, the effect is more pronounced for motion blur than for Gaussian blur. This is likely because motion blur rarely happens naturally in still images, and so it is almost entirely absent in testVOC, making the equalization process distort the training set further away from the test set statistics. This also reveals that motion blur is a more important domain difference between VOC and YTO than Gaussian blur.
Testing on YTO shows an interesting phenomenon: using blurred training samples has a much smaller effect. This 
makes sense as testYTO is already naturally blurred, and therefore blurring the training set does not lose much relevant information.

Equalizing image quality with Gaussian blur reduces the performance gap when testing on VOC down to $7.0\%$ mAP for  DPM and $8.1\%$ for R-CNN. Motion blur makes the gap even smaller, to $5.1\%$ for DPM and $6.3\%$ for R-CNN. The amount of gap bridged for R-CNN is remarkably large ($8.7\%$ mAP).
When testing on YTO, Gaussian blur leaves the gap essentially unchanged for both detectors, while motion blur widens the gap for R-CNN by a small amount of $2.7\%$, reaching $8.4\%$ mAP.
%
Given that motion blur better represents the relevant image quality difference between the two domains, in the following we work only with motion blurred training sets.

\vspace{-2mm}
\subsection{Aspect distribution}
\label{sec:aspects}

\begin{figure}[t]
\begin{center}
\includegraphics[width=\linewidth]{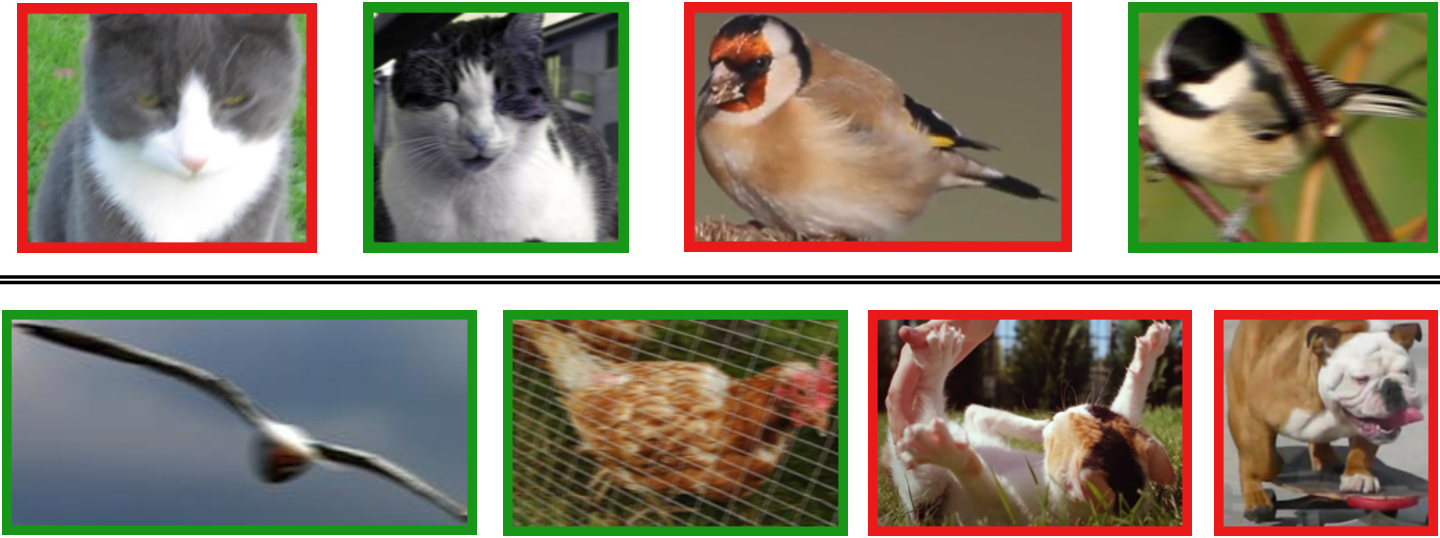}
\vspace*{-8mm}
\caption{\textit{Top row: Aspects common to both VOC (green) and YTO (red) domains.
Bottom row: Red: aspects occurring only in YTO. YouTube users often film their own pets doing funny things, such as jumping, rolling and going on a skateboard. Green: aspects occurring only in VOC. Chicken and birds flying are common in VOC, but do not appear in YTO.}}
\vspace{-5mm}
\label{figure:more_aspects}
\end{center}
\vspace{-2mm}
\end{figure}

As the last factor, we consider the distribution over aspects, i.e.
the type of object samples in the training sets.
Differences can be due to biases in the distribution of 
viewpoints, subclasses, articulation and occlusion patterns. 
As the space of possible samples for an object class is very large, any given dataset invariably
samples it in a limited way, with its own specific
bias~\cite{torralba2011cvpr}.
Fig.~\ref{figure:horse_kl} illustrates this point by showing all training samples of the class
`horse' from both domains. The distributions differ considerably and
overlap only partially.
Horses jumping over hurdles appear in trainVOC but not in trainYTO, while the latter has more horses running free in the countryside (more examples in fig.~\ref{figure:more_aspects}).
We work here with the most equalized training sets, i.e. trainVOC Motion Blurred Unique Samples and trainYTO
Unique Samples. These have the same size, accuracy of spatial
support, level of appearance diversity, and image quality.

\vspace{-1mm}
\paragraph*{Measurement} 
\phantomsection
\addcontentsline{toc}{section}{Measurement}
\label{As_M}

We refer to the two training sets as $A = \left( x^{A}_{1} , \dots , x^{A}_{n} \right)$ for VOC, and $B = \left( x^{B}_{1} , \dots , x^{B}_{n} \right)$ for YTO.
We measure the difference in their aspect distributions with the symmetrized Kullback-Leibler (KL) divergence 

\begin{footnotesize}
\[
d_{KL} \left(A, B \right)
 = \sum \limits_{ i=1 }^n \left( \hat{f_{A}} \left( x^{A}_{i} \right) \ln \frac{\hat{f_{A}} \left(  x^{A}_{i} \right) }{ \hat{f_{B}} \left( x^{B}_{i} \right) } + \hat{f_{B}} \left( x^{B}_{i} \right) \ln \frac{ \hat{f_{B}} \left( x^{B}_{i} \right) }{\hat{f_{A}} \left( x^{A}_{i} \right) } \right)
\]
\end{footnotesize}
\noindent where 
\begin{small}
\[\hat{f_{s}}\left( x^{s}\right) = \frac{1}{n} \sum \limits_{ i = 1} ^{ n } K \left( x^{s}-x^{s}_{i} ; h \right),  \quad s \in \{A,B\}
\]
\end{small}

\noindent is a kernel density estimator fit to sample set $s$; $K \left( \:\cdot \: ; h\right) $ is the isotropic Gaussian kernel with standard deviation $h$ (automatically set based on the standard deviation of the sample set~\cite{matlab-kde}).
%
For ease of visualization and computational efficiency, we reduce the dimensionality of the CNN features to 2D, using the algorithm of \cite{Maaten08machine} (as done by~\cite{donahue13decaf,jia2014caffe}).

The KL divergence between the two training sets, averaged over all classes, is $5.25$. This shows that the difference in aspect distribution is quite big, given that the KL divergence averaged over all classes between trainVOC and testVOC is $1.24$.

\vspace{-1mm}
\paragraph*{Equalization} 
\phantomsection
\addcontentsline{toc}{section}{Equalization}
\label{As_E}

We equalize the aspect distributions by subsampling the two training sets such that the subsets 
have a similar aspect distribution, i.e. a small $d_{KL}$.
More precisely, we want to find the largest subsets $\widetilde{A} \subset A$ and $\widetilde{B} \subset B$ which have a small enough $d_{KL}$ to be considered equally distributed: $d_{KL}(\widetilde{A}, \widetilde{B}) < \epsilon$. We approximate this optimization by a greedy forward selection algorithm that starts from $\widetilde{A} = \widetilde{B} = \emptyset$, and iteratively adds the pairs of samples with the smallest Euclidean distance. We stop growing the subsets when $d_{KL}$ exceeds $\epsilon$.

Fig.~\ref{figure:horse_kl}~(right) illustrates the evolution of $d_{KL}(\widetilde{A}, \widetilde{B})$ during this process for the class `horse'. 
We use a small $\epsilon = 0.1$ in all experiments \footnote{Note the need for this parameter, otherwise $d_{KL}(\widetilde{A}, \widetilde{B})=0$ is achieved by picking $0$ samples from each set.}. 
For the horse class, this selects $70$ samples from each set, which is just after the horizontal portion of the curve in fig.~\ref{figure:horse_kl}, when the distributions start to differ significantly, see samples along the curve. 
Fig.~\ref{figure:horse_kl}(left) depicts each a selected pair of samples, which lies in the region where the distributions overlap. 
This pair show samples with similar aspects, whereas distant samples typically show very different aspects. 

This procedure constructs the new training sets ``trainVOC Motion Blurred Unique Samples and Aspects'' and ``trainYTO Unique Samples and Aspects''. These sets contain $551$ samples each (about $1/4$ of all samples left after equalizing the previous domain shift factors).

\vspace{-1mm}
\paragraph*{Impact} 
\phantomsection
\addcontentsline{toc}{section}{Impact}
\label{As_I}

We train object detectors from the aspect-distribution equalized sample sets. Results for R-CNN on both test sets are reported in fig.~\ref{figure:map}c,d (we do not perform this equalization process for DPM, as it does not work on a simple feature space where we can easily measure distances, due to its movable parts).
When testing on VOC, the performance of training from YTO is barely affected by the equalization process, despite having $4\times$ fewer training samples. Instead, the mAP of training from VOC drops considerably. This can be explained by the fact that the equalization process returns the intersection of the distributions of the two training sets. The video training set only loses samples with aspects not occurring in the trainVOC distribution, and hence unlikely to be in the test set. 
Instead, the VOC training set is losing many aspects that do occur in the test set.
Testing on YTO corroborates this interpretation by displaying the inverse behaviour: the performance of training on VOC remains essentially unchanged, whereas that of training on YTO worsens substantially. 

Equalizing the aspect distributions brings the performance of when testing on VOC down to just $2.0\%$ mAP, closing it by $4.3\%$. When testing on YTO, the equalization has an even greater effect: it bridges the gap by $6.9\%$ mAP, reducing to just $1.5\%$.

The results show that aspects play an important role. Performance depends considerably on the training set containing aspects appearing in the test set, and this matters more than the size of the training set. The results also shows that the aspect distributions in trainVOC and trainYTO are quite different, a dataset bias phenomenon analog to that observed by~\cite{torralba2011cvpr} when studying the differences between still image datasets. 
Our findings provide a guideline for practitioners trying to enrich still image training sets with video data (or vice-versa): it is more important to carefully consider {\em which} data samples to add, rather than simply trying to add a large number of them.

\vspace{-3mm}
\subsection{Other factors}
\label{sec:other_factors}

In addition to the four domain shift factors we studied in sec.~\ref{sub:SLA} to ~\ref{sub:AD}, we also considered other factors, which we summarize here. However, when measuring these other factors, we did not observe any significant differences between the VOC and YTO, and so we did not proceed to the equalization and impact steps. 

\vspace{-1mm}
\paragraph*{Object size and aspect-ratio}
The average size of ground-truth bounding-boxes in trainVOC, relative to the size of the image, is $0.26$. For trainYTO it is nearly the same ($0.25$). Similarly, the average aspect-ratio (width-over-height) is $1.48$ for trainVOC vs $1.53$ for trainYTO.

\vspace{-1mm}
\paragraph*{Camera framing}
We look for differences in the way objects are framed by the camera. Potentially, YTO might have more objects coming in and out of the frame. Each VOC instance is annotated by a tag marking it as either normal, truncated (partially out of the image frame),
or difficult (very small, very dark, or heavily occluded)~\cite{pascal07:thomas}.
In order to measure camera framing for trainYTO, we annotated all its instances with the same tags.
Both trainVOC and trainYTO have about the same proportion of truncated instances ($35.8\%$ in trainVOC, $33.2\%$ in trainYTO).
We exclude instances marked as difficult from trainVOC, as they are not taken into account in the PASCAL VOC 2007 either.
Only $0.3\%$ of the trainYTO set are difficult instances, again leading to about the same percentage.
All other instances are normal (i.e. about $65\%$ in both trainVOC and trainYTO).


\vspace{-4mm}
\section{Experiments on ILSVRC 2015}
\label{sec:ILSCVRC}

\begin{figure}[t]
\begin{center}
\includegraphics[width=\linewidth]{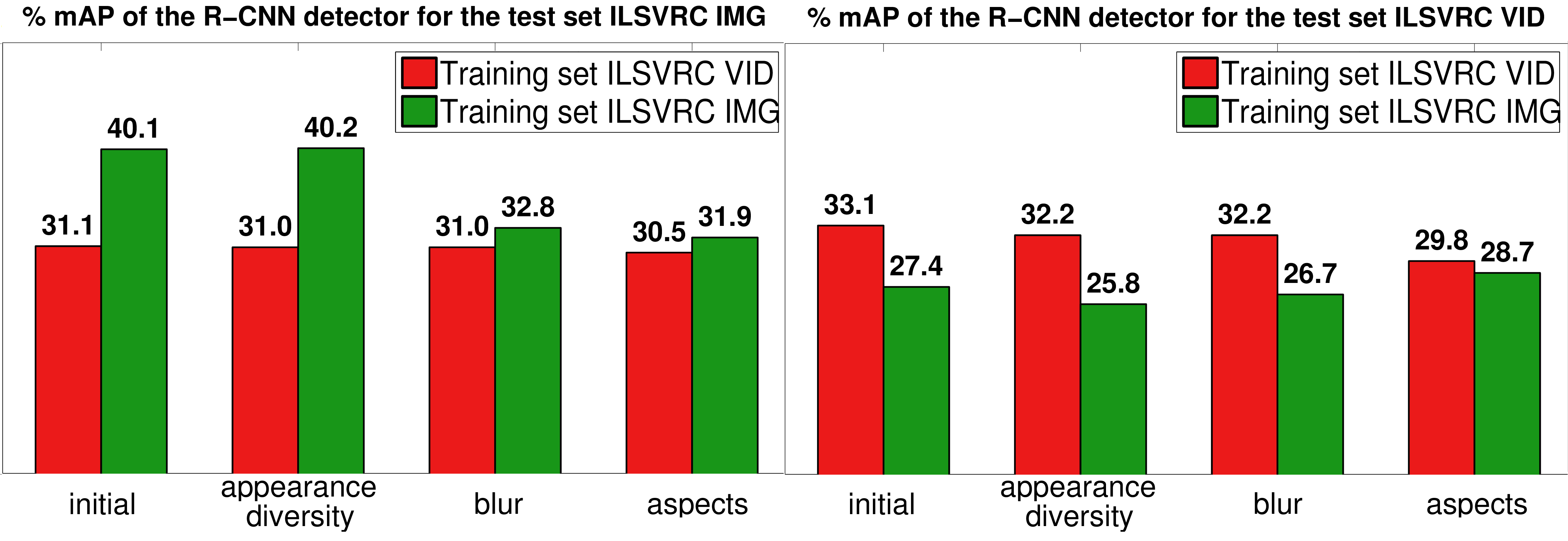}
\vspace*{-6mm}
\caption{\textit{Impact of the domain shift factors when training on ILSVRC IMG and VID
  for the R-CNN detector.}}
\vspace{-4mm}
\label{figure:map2}
\end{center}
\vspace{-3mm}
\end{figure}

\ifx
\begin{figure}[t]
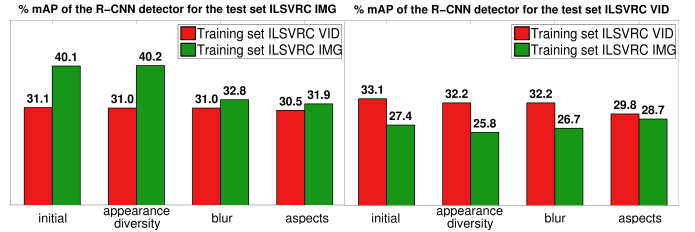

\centerline{%
\begin{tabular}{c@{}c@{}c@{}c@{}c@{}}
\includegraphics[width=0.9\linewidth]{newfigs/R-CNN_ILSVRC IMGFin.pdf} \\
\includegraphics[width=0.9\linewidth]{newfigs/R-CNN_ILSVRC VIDFin.pdf} \\
\end{tabular}}
\vspace*{-3mm}
\caption{\textit{Impact of the domain shift factors when training on ILSVRC IMG and VID
  for the R-CNN detector.}}
\vspace{-4mm}
\label{figure:map2}
\end{figure} 
\fi

We repeat here our analysis on another dataset pair, to verify that our findings are general.
Both datasets in this pair come from the ImageNet Large Scale Visual Recognition Challenge (ILSVRC
\cite{ILSVRC15,ImgNet2015}), i.e. from the object detection in images (IMG) and in videos (VID) tracks of the challenge. We consider the same 10 object classes as in the previous sections.

\paragraph*{Data and protocol}
The ILSVRC IMG dataset contains 60k training images (train60k) and 20k validation images, fully annotated with bounding-boxes on all instances of $200$ object classes. We split the validation set into val1 and val2 as in~\cite{girshick14cvpr}. For training, we use train60k+val1, resulting in $13,335$ bounding-boxes for our 10 classes ($8,021$ images).  For testing we use val2, which comprises $5,310$ bounding-boxes ($3,362$ images). 
The ILSVRC VID dataset contains $3,862$ training and $555$ validation video snippets, which we use for training and testing respectively. The snippets are manually annotated with bounding-boxes for $30$ object classes.
For our $10$ classes, the training set has $2,198$ snippets, totalling $292,199$ bounding-boxes in $212,643$ frames.
The validation set, used as test set, has $332$ snippets, with $134,432$ bounding-boxes in $87,715$ frames. 

We apply the equalization procedure of sec.~\ref{sub:da} to have the same number of training samples in each domain.
This results in $13,335$ training samples and $3,362$ test images per domain. 
We refer to the two training sets as trainIMG and trainVID. 
Following the protocol of sec.~\ref{Protocol}, we train an R-CNN object detector either from still images or from video frames, then test it on both domains, and finally measure performance by mAP on the test set. 

\paragraph*{Domain shift factors}
We analyze $3$ out of the $4$ domain shift factors from sec.~\ref{sec:factors}. We do not examine the spatial location accuracy factor, since we start from perfect spatial support (ground-truth bounding-boxes).
For the appearance diversity factor, $96.6\%$ of the samples in trainIMG are unique, whereas only $61.6$\% trainVID samples are unique, analog to what observed on the PASCAL VOC - YouTube-Objects dataset pair. 
We apply the equalization procedure of sec.~\ref{sub:AD}, obtaining two new training sets, each containing $7,902$ unique samples. 
For image quality, the gradient energy averaged over all classes is $4.4$ for the unique samples in IMG (identical to VOC) and $3.0$ for those in VID (so, blurrier than YTO).
By applying the Gaussian blur filter of sec.~\ref{sub:IQ} on the image samples, we equalize their blur level to match the VID samples.
For aspect distribution, the KL divergence between the two training sets, averaged over all classes, is 
$7.52$. We apply the aspect distribution equalization procedure of sec. \ref{sec:aspects}, resulting in the two final training sets, each containing $3,446$ samples.


Fig.~\ref{figure:map2} shows the evolution of the performance of the R-CNN detector on the test sets after canceling out each domain shift factor in turn.
Generally, we observe the same trend as in the YTO-VOC pair, i.e. the gap is initially rather substantial, and it is gradually reduced by our equalization steps. The final gap after all steps is below $1.5\%$ mAP on both test sets. 

When looking closer, some differences to the YTO-VOC results appear.
When testing on images, the appearance diversity factor leaves the gap unchanged. This is due to the larger number of training samples in ILSVRC IMG, compared to VOC ($4\times$ more). Even after removing about $40\%$ of the unique training samples from ILSVRC IMG in order to match the number of unique samples in ILSVRC VID, there are still enough samples left to train good detectors.
Interestingly, when testing on images, the image quality factor closes the gap by a large margin. This is due to ILSVRC VID being blurrier than YTO, so the image quality equalization applies a stronger blur to ILSVRC IMG than to VOC.
The aspect distribution factor bridges the performance gaps for both domains, in line with what observed on VOC-YTO.
This confirms the important impact that the aspects contained in a training set have on performance at test time.

\ifx
\section{bag}
\label{bag}

Choice of our 3 factors:

1. spatial location accuracy: as most previous experiments on training detectors from video were done in a weakly supervised setting, it was natural to assume that much of the performance gap was due to the poor quality of automatically generated bounding-boxes. Instead, our experiments with ground-truth bounding-boxes show that even great future progress on video segmentation and weakly supervised learning will not solve the problem.

2. appearance diversity: video differs from still images as frames are temporally correlated. This is an intrinsic difference in the medium, and leads to this often overlooked factor. Our experiments suggest to work with many shots, sampling just a few frames from each.

3. image quality: this is a more apparent difference between video and still images. Our experiments show that it does play a significant role, but it is not the only culprit.

We said that we have to comment at some point that we generally compare 1 instance per domain, motivating the image and video domains..

Also, we said that we may want to mention that we don't really present results for images/ videos generally but for voc and yt. maybe mention that they are representative?

\fi

\vspace{-4mm}
\section{Conclusions}
\label{sec:conclusions}

We analyzed several domain shift factors between still images and video frames for object detection. This is the first study that addresses with a systematic experimental protocol such an important task. We believe our conclusions are valuable in promoting and guiding future research.
We thoroughly explored $4$ domain shift factors and their impact on the performance of two modern object detectors~\cite{felzenszwalb10pami,girshick14cvpr}.
We showed that by progressively cancelling out these factors we gradually closed the performance gap between training on the test domain and training on the other domain.

Given that data is becoming abundant, it is important to decide {\em which} data to annotate so as to create better  object detectors.
Our experiments lead to several useful findings, especially relevant when trying to train detectors from video to perform well on image test sets:
(1) training from video with ground-truth bounding-box annotation still produces a worse detector than when training from still images. Hence, future research on video segmentation cannot solve problem \textit{on its own};
(2) blur has a strong impact on the performance gap; hence, deblurring algorithms might be an avenue for removing this factor; 
(3) the appearance diversity and aspect distribution of a training set is much more important than the number of training samples it contains. 
For good performance one should collect a broad range of videos showing all aspects expected to appear in the test set.

\ifx

We have analyzed several domain shift factors between still images and video frames for object detection. This is the first study that addresses with a systematic experimental protocol such an important task. We believe our conclusions are valuable in promoting and guiding future research.
We have thoroughly analysed four domain shift factors and their impact on the performance of two modern object detectors~\cite{felzenszwalb10pami,girshick14cvpr}.
We showed that by progressively cancelling out these factors we gradually closed the performance gap between training on the test domain and training on the other domain.

Given that data is becoming abundant, it is important to decide {\em which} data to annotate so as to create better  object detectors.
Our experiments lead to several useful findings, especially relevant when trying to train detectors from video to perform well on still image test sets:
(1) training from video with ground-truth bounding-box annotation still produces a worse detector than when training from still images. Hence, future research on video segmentation cannot solve problem \textit{on its own};
(2) motion blur is a stronger cause of performance gap than Gaussian blur. Hence, training videos with a slow moving camera and/or objects are preferable;
(3) the appearance diversity and aspect distribution of a training set is much more important than the number of training samples it contains. For good performance one should collect a broad range of videos showing all aspects expected to appear in the test set.

\fi

\ifCLASSOPTIONcompsoc
  \vspace{-3mm}
  \section*{Acknowledgments}
\else
  \section*{Acknowledgment}
\fi

We gratefully acknowledge the ERC projects VisCul and ALLEGRO.

\ifCLASSOPTIONcaptionsoff
  \newpage
\fi



%
\vspace{-4mm}
{\small
\bibliographystyle{ieee}
\bibliography{shortstrings,vicky_iccv}
}



%








\end{document}